# Brain-Inspired Continual Learning: Robust Feature Distillation and Re-Consolidation for Class Incremental Learning

Hikmat Khan
*Rowan University*

Nidhal Carla Bouaynaya
*Rowan University*

Ghulam Rasool

**RESEARCH ARTICLE**

# Brain-Inspired Continual Learning: Robust Feature Distillation and Re-Consolidation for Class Incremental Learning

HIKMAT KHAN[1], NIDHAL CARLA BOUAYNAYA[1], (Member, IEEE), AND GHULAM RASOOL[2], (Member, IEEE)
[1]Department of Electrical and Computer Engineering, Rowan University, Glassboro, NJ 08028, USA
[2]Department of Machine Learning, Moffitt Cancer Center, Tampa, FL 33612, USA

Corresponding author: Nidhal Carla Bouaynaya (bouaynaya@rowan.edu)

This work was supported in part by the National Science Foundation under Award ECCS-1903466 and Award OAC-2008690. The work of Hikmat Khan was supported by the Cooperative Agreement under Grant 16-G-015.

**ABSTRACT** Artificial intelligence and neuroscience have a long and intertwined history. Advancements in neuroscience research have significantly influenced the development of artificial intelligence systems that have the potential to retain knowledge akin to humans. Building upon foundational insights from neuroscience and existing research in adversarial and continual learning fields, we introduce a novel framework that comprises two key concepts: feature distillation and re-consolidation. The framework distills continual learning (CL) robust features and rehearses them while learning the next task, aiming to replicate the mammalian brain's process of consolidating memories through rehearsing the distilled version of the waking experiences. Furthermore, the proposed framework emulates the mammalian brain's mechanism of memory re-consolidation, where novel experiences influence the assimilation of previous experiences via feature re-consolidation. This process incorporates the new understanding of the CL model after learning the current task into the CL-robust samples of the previous task(s) to mitigate catastrophic forgetting. The proposed framework, called Robust Rehearsal, circumvents the limitations of existing CL frameworks that rely on the availability of pre-trained Oracle CL models to pre-distill CL-robustified datasets for training subsequent CL models. We conducted extensive experiments on three datasets, CIFAR10, CIFAR100, and real-world helicopter attitude datasets, demonstrating that CL models trained using Robust Rehearsal outperform their counterparts' baseline methods. In addition, we conducted a series of experiments to assess the impact of changing memory sizes and the number of tasks, demonstrating that the baseline methods employing robust rehearsal outperform other methods trained without robust rehearsal. Lastly, to shed light on the existence of diverse features, we explore the effects of various optimization training objectives within the realms of joint, continual, and adversarial learning on feature learning in deep neural networks. Our findings indicate that the optimization objective dictates feature learning, which plays a vital role in model performance. Such observation further emphasizes the importance of rehearsing the CL-robust samples in alleviating catastrophic forgetting. In light of our experiments, closely following neuroscience insights can contribute to developing CL approaches to mitigate the long-standing challenge of catastrophic forgetting.

## I. INTRODUCTION

Continual Learning (CL), also referred to as incremental, lifelong, or sequential learning, equips deep learning models with the ability to accumulate and expand knowledge over time, similar to humans [1], [2], [3]. Despite advancements in CL methodologies, current approaches still suffer from a phenomenon known as *catastrophic forgetting* or *catastrophic interference* [1], [4], which predominantly occurs when the model trained on a sequence of tasks exhibits degraded performance on the earlier tasks after learning a new task [4], [5]. This challenge necessitates the development of approaches that enable models to learn continuously without compromising performance on previously learned tasks. Several approaches have been proposed to mitigate catastrophic forgetting in CL. *Regularization-based approaches* focus on penalizing changes to the model's weights or features crucial for previous tasks. The intent is to maintain the model's proximity to its former states or representations, thereby preserving existing knowledge while learning the new task [6], [7], [8], [9], [10], [11]. Regularization approaches face the critical challenge of maintaining an equilibrium between the retention of acquired knowledge and the plasticity to adapt to new tasks, a balance often resulting in suboptimal performance in both aspects [5]. *Dynamic architecture-based approaches* involve dynamically adding or removing neurons, layers, or dedicated sub-networks, aiming to increase the model's capacity to learn new tasks while preserving the existing layers or sub-networks unchanged [12], [13], [14], [15], [16], [17], [18]. However, such adaptability may lead to increased complexity and challenges in maintaining architectural stability over time [12]. *Replay-based methods* provide the model access to a subset of previous data by storing them in the rehearsal or replay memory [18]. Interleaving these previously stored samples from the replay memory with the current mini-batch enables the model to retain knowledge from earlier tasks [19], [20], [21], [22], [23], [24], [25], [26]. However, a challenge persists in determining the optimal selection and quantity of old samples for rehearsal to achieve optimal performance, thus complicating the effective management of the replay memory [22]. *Knowledge distillation-based methods* minimize the divergence between previous features or representations used to solve previous tasks and the new features or representations that emerge as the model learns new tasks [7], [27], [28], [29], [30]. *Dataset distillation-based methods*, motivated by the concept of dataset distillation as introduced in [31], [32], [33], [34], [35], [36], and [37], involve either pre-distilling the entire dataset or learning retrievable memories for previous tasks, replayed while the model learns new tasks. However, challenges emerge when pre-distilling the entire dataset or static features from prior tasks and then incorporating them into sequential learning adversely affect the model's performance. This issue arises from the necessity to accommodate time-varying features for prior tasks after the CL model assimilated the new task [38], [39]. To overcome such limitation, our previous work [39] proposed using a pre-trained Oracle CL model for distilling a CL-robustified dataset for each task, subsequently utilized for training the final CL model. However, the assumptions about the availability of a pre-trained CL model and pre-distilled CL-robustified datasets restrict the practical feasibility of the proposed method [39]. *Robust feature distillation-based methods* aim to assess the impact of the features provided to CL models during the training on the robustness and catastrophic forgetting of the final trained CL model. In our previous work, [40], we demonstrated that features hold an equally pivotal role in both CL and adversarial domains. Specifically, we demonstrated in [40] that the CL models trained on adversarially-robust features exhibit a superior ability to consistently maintain CL performance under varying natural and adversarial noise, whereas the CL models trained on non-robust features suffer performance degradation [40], [41]. Furthermore, to facilitate the CL model's acquisition of adversarially robust features, the authors in [42] incorporated adversarial training for each task. The adversarial training improved the CL model's robustness and performance under various adversarial conditions [42], further underscoring the vital role of available features in influencing the CL model's performance. *Brain-inspired methods* draw inspiration from neuroscience, aiming to emulate biological principles in computational techniques to improve memory retention in CL models [43], [44], [45], [46], [47]. *Hybrid methods* integrate diverse CL learning strategies to capitalize on their strengths, thereby proposing models that exhibit reduced susceptibility to catastrophic forgetting [3].

This paper builds upon prior contributions to continual learning and adversarial learning and incorporates a series of crucial observations derived from neuroscience. Deep learning models, especially neural networks, are designed to autonomously learn features from raw input data without differentiation, driven primarily by the objective of minimizing the loss function. As highlighted by the authors in [48], input data comprises both robust and non-robust features. Robust features encapsulate attributes specific to individual classes, facilitating the model's ability to make robust classifications. In contrast, non-robust features are distributed all across various classes and are characterized by properties that enhance a model's generalization capabilities, albeit at the expense of model robustness [48]. A model trained on a dataset comprising both robust and non-robust features tends to prioritize learning non-robust features due to their generalizability, which facilitates minimizing the loss function at the expense of model robustness. In light of these observations, the authors in [48] proposed a framework to discern between the robustified and non-robustified versions of the standard input dataset. The robustified version encapsulates robust features, while the non-robustified version predominantly comprises non-robust features. Deep learning models trained on the robustified version showed greater robustness to adversarial attacks than those trained on non-robustified features [48]. These observations underscore features' pivotal influence in shaping the trained models' overall characteristics. In a similar context, our prior work in [40] extended such observations to the domain of CL, and similar and consistent findings were observed [40], illustrating that CL models trained on robust features performed robustly under noisy and adversarial conditions, in contrast to the CL models trained on non-robust features.

These works highlight the critical importance of the features present in the input and extracted by the model, impacting performance in both adversarial and continual learning settings [40], [48]. Recently, we expanded the adversarially-robust framework introduced in [48] to the continual learning setting by generating CL-robustified versions of the input dataset using a pre-trained CL Oracle model. This framework operates under a similar assumption to that employed in [48], relying on the availability of a pre-trained Oracle CL model for distilling the refined CL-robustified dataset. Consistent with [48], our observations revealed that the CL model trained on CL-robustified datasets suffered less from catastrophic forgetting than counterparts trained on the standard dataset. These observations in [39] underscore that a CL model trained on a pre-distilled CL-robustified dataset mitigates catastrophic forgetting, emphasizing the capacity of CL-robustified features in mitigating catastrophic forgetting.

Although our prior research conducted in [39] underscores the pivotal role of distilled CL-robustified features in CL models, it is noteworthy that they introduced two practical limitations: First, they assumed the availability of the pre-trained Oracle CL model to use in the distillation of the robustified CL features. Second, it presumed the existence of pre-distilled CL-robust features used for training subsequent CL models. These assumptions are impractical, as they entail key preconditions: the use of a pre-trained Oracle model to distill CL-robust features for subsequent CL models to mitigate catastrophic forgetting.

To address the identified limitations, this paper introduces feature distillation and re-consolidation inspired by neuroscience to emphasize that replaying CL-distilled and re-consolidated features is crucial for preserving previous knowledge [43], [44], [46], [49]. The first neuroscience inspiration relates to how the mammalian brain replays the waking experience in the distilled and compressed form (i.e., not in the original form and time scale) to perform memory consolidation [50], [51], [52], [53], [54], [55], [56]. Furthermore, It has been shown that the brain rehearses past experiences during the post-learning, sleeping phase to consolidate memories [57], [58], [59], [60], [61]. These observations from neuroscience highlight the importance of rehearsal-based strategies and replaying the distilled version of the waking experiences [50]. Taking inspiration from such observations, we introduce feature distillation to distill CL-robustified features for memory samples, which are then replayed during the learning of subsequent tasks using a rehearsal-based strategy.

Second, novel waking experiences have a pronounced impact on previous experiences [49], [50], [62]. These observations emphasize the necessity to update the CL-robustified features of previous tasks during the learning of the current task to incorporate essential modifications and update the relative understanding of earlier tasks. Therefore, to closely replicate similar steps in CL, we introduce the feature re-consolidation step to re-distill the features of the previous tasks. In addition to drawing insights from neuroscience, feature re-consolidation has also been motivated by the cycling method proposed in prior studies [63], [64] to enhance the alignment of features of previous tasks after the model learns the current task. The feature re-consolidation is performed for prior tasks to re-distill the updated knowledge of previously learned tasks [49], [50], [62] into the CL-robustified memory samples.

Third, the mammalian brain does not reenact waking experiences in their original form and instead replays them at a significantly accelerated time scale [49], [65], [66], [67], [68]. This observation underscores the importance of replaying distilled or CL-robustified samples containing the sparse CL-robustified features, see Figure 6, while learning the following tasks.

In this paper, First, to shed light on the existence of diverse features, we pose the research inquiry of whether various training objectives, including multi-task learning, continual learning, or adversarial learning, induce the model to uncover diverse features. Second, we propose a neuroscience-inspired framework that addresses the limitation identified in [39]. The framework approach consists of two neuroscience-inspired modifications: First, feature distillation entails the distillation of CL-robustified features pertinent to the current task. Second, feature re-consolidation focuses on re-distilling the CL-robustified features, thereby enabling the incorporation of updated feature importance information for previous tasks after the model learns the current task [49], [65], [66], [67], [68]. The feature re-consolidation ensures recalibration of CL-robust features associated with previous tasks, thus accommodating the evolving dynamics of CL-robust features [49], [50], [62]. The proposed framework is based on a replay-based strategy called Robust Rehearsal. Our experimental results demonstrate that CL models trained using robust rehearsal outperform the CL baselines outlined in [39] and obviate the necessity for a pre-trained Oracle CL model and pre-distilled CL-robustified dataset. In addition, we conducted a series of experiments across varying sizes of the rehearsal memory to demonstrate the efficacy of the feature distillation and re-consolidation steps in mitigating catastrophic forgetting compared to a naive rehearsal-based CL strategy.

The main contributions of this paper are as follows:

- Taking inspiration from the working principles of the brain, we introduce a novel framework called "Robust Rehearsal." This framework initially distills CL-robust features and then performs re-distillation and

- re-consolidation to assimilate revised interpretations of prior task knowledge. Our extensive experiments show that robust rehearsal is crucial in reducing catastrophic forgetting, outperforming baseline methods trained without robust rehearsal.
- We extended upon our prior work in [39] by introducing three levels of distillation loss: input, feature, and prediction space. The input component of the loss is inspired by neuroscience findings that novel waking experiences influence previous experiences [49], [50], [62]. The proposed distillation aims to align the distilled CL-robust samples visually more closely with the original class's feature [49], [50], [62]. Interestingly, the resulting distilled CL-robust samples exhibit characteristics reminiscent of human dream imagery. These observations draw parallels with the phenomenon in the biological brain, where waking experiences are recapitulated in a distilled form and at an accelerated timescale for memory consolidation.
- The proposed distillation loss, consisting of feature distillation and re-consolidation, mitigates the limitations of previous frameworks that relied on pre-trained Oracle CL models and pre-distilled robustified datasets for CL.
- The proposed approach aligns with existing rehearsal-based CL approaches and can readily enhance their performance on standard CL tasks.
- Lastly, we empirically demonstrated the impact of different training objectives, including joint, continual, and adversarial learning, on discovering diverse features. The findings lead to the insightful conclusion that optimization objectives play a substantial role in governing the feature learning process within deep learning models, underscoring the importance of optimal feature acquisition in various learning paradigms

The remainder of this paper is organized as follows: Section II presents a review of related work. Section III outlines the problem setup and motivation, with comprehensive details on feature distillation and re-consolidation losses. Section III also provides details on the proposed robust rehearsal and the experimental setup. Section IV-B presents results and discussion. Section IV-C highlights the limitations of the proposed methodology and proposes potential avenues for future research. Finally, Section V summarizes key findings and concludes the paper.

## II. RELATED WORK

Existing CL approaches can be roughly divided into the following categories: 1) regularization-based approaches, 2) architecture-based approaches, 3) rehearsal-based approaches, 4) knowledge distillation-based approaches, 5) dataset distillation-based approaches, 6) feature distillation-based approaches, 7) Brain-inspired approaches and 8) Hybrid approaches. **Regularization-based** approaches deter updates to weights or parameters vital to the performance of the already learned tasks by incorporating an additional regularization term into the model's loss function [6], [7], [8], [9], [10], [11], [69]. The regularization term restricts the updates (or plasticity) of parameters crucial for old tasks to ensure that the model's weights or features remain proximal to their former states or representations, thereby conserving knowledge of old tasks as the model learns new tasks [9]. **Dynamic architecture-based** approaches dynamically expand the network or use masks to activate a sub-network or selectively connect sub-networks in the existing architecture to adapt to new tasks [12], [13], [14], [15], [16], [17], [18]. More specifically, these methods expand the network by incorporating task-specific sub-networks, each with its unique set of parameters, for every new task [14], [70], [71], [72], [73], [74], [75]. **Rehearsal-based** approaches predominantly leverage a rehearsal buffer or reply memory to preserve a subset of data samples from previously learned tasks, either stored and replayed in their original raw form or replayed in the pseudo-generated form [18], [19], [20], [21], [22], [23], [24], [25], [26], [44], [64], [76], [77], [78], [79], and [80]. The buffered or rehearsal samples are then interleaved with the current task data, providing the model limited access to the previous task data while learning new tasks. The introduction of additional distillation penalties further enhances the rehearsal-based approaches [19], [23], [27], [81], [82], [83], resulting in state-of-the-art performance across various benchmark datasets [2], [84]. **Knowledge distillation-based** approaches aim to minimize the divergence from representations of the previous model while effectively adapting to new tasks [7], [27], [28]. These approaches integrate the principles of knowledge distillation, initially introduced in [85] and [86], which involves the transfer of knowledge from a large model to a smaller, deployable and more efficient model while maintaining its performance closely aligned with that of the large model. In the context of CL, the previous task's model, at $t-1$, is retained and leveraged to distill essential feature knowledge of past tasks, which is subsequently incorporated into the current model's learning process, at time $t$, to ensure retention of previous knowledge while learning the new task. **Brain-inspired** approaches draw inspiration from neuroscience, particularly from the observation that the biological brain rehearses only the most crucial aspects of pre-sleep experiences in dreams to consolidate memory [49]. Additionally, the rehearsal of waking experiences occurs in intermittent bursts, not in the original form, and on a faster timescale than the original experience, in brain [49], [53], [54], [56], [65], [87]. The CL approaches, inspired by neuroscience, strive to closely translate the insights from the understanding of the biological brain into practical strategies to mitigate catastrophic forgetting [43], [44], [45], [46], [47], [88], [89], [90], [91], [92], [93], [94]. The authors in [95] introduced a universal sleep decoder to align neural representations between wakefulness and sleep. The authors in [96] drew inspiration from the hippocampus's role in the human brain, notably its function in transitioning short-term memory to long-term memory, to implement

Artificial Hippocampi for lifelong learning. The authors in [97] emphasized replay-based echoing biological memory mechanisms, while the biological brain engages in selective memory retention of salient experiences, and introduced a framework that combines associative memory with replay-based strategies, thereby bringing CL closer to human memory processes. The authors in [98] drew inspiration from biological nervous system mechanisms, introducing time-aware regularization to dynamically fine-tune the generator for synthesizing pseudo-data points of previously learned tasks, which are subsequently replayed alongside new tasks during concurrent training. **Dataset distillation-based** approaches, motivated by the concept of dataset distillation as introduced in [31], [32], [33], [34], [35], [36], and [37], involve the distillation of knowledge from large datasets into a reduced number of synthesized samples, such that the same model, trained on the original large dataset and its distilled version, achieves comparable performance. In the context of CL, such approaches based on dataset distillation involve either pre-distilling the entire dataset or learning retrievable memories for previous tasks, which are subsequently replayed during the model's adaptation to new tasks [38], [39], [99]. Specifically, the approach described in [38] involved learning a set of bases shared across various classes, referred to as memories. These shared memories are then combined through learned functions to generate synthesized replay samples for previously learned tasks while the model learns a new task. However, it has been empirically observed that the pre-distillation of the entire dataset or freezing of features from prior tasks, and their incorporation into sequential learning, adversely affect the model's performance [38], [39]. This limitation emerges due to the sequential nature of learning in CL models, which necessitates the dynamic adjustment of the relevance of previously acquired features for earlier tasks during incremental learning. To mitigate this challenge, our study presented in [39] employs pre-trained Oracle CL models specific to each task to construct a CL-robust dataset for each corresponding task. Subsequently, the refined CL-robust dataset was employed to train future CL models. Nonetheless, the presupposition regarding the availability of both the pre-trained Oracle model and the pre-distilled CL-robust dataset when training final CL models constrains the practical applicability of the proposed methodology [39]. Recently, **Robust Feature Distillation** approaches motivated by dataset distillation aim to assess the impact of the features provided to CL models on the robustness and catastrophic forgetting of CL models. It has been demonstrated that features play a pivotal role in both the CL and adversarial domains [40], [48]. Specifically, as shown in [40], CL models trained on adversarially-robust features exhibit a superior ability to maintain consistent CL performance under adversarial conditions, whereas CL models trained on non-robust features suffer performance degradation [40], [41]. To facilitate the CL model's effective acquisition of adversarially robust features, the authors in [42] incorporated adversarial training for each task. The adversarial training led to a notable enhancement in the CL model's robustness and performance across a spectrum of adversarial conditions [42], which emphasized the crucial role played by the available features in influencing the performance of the CL model. **Hybrid** approaches combine and aim to leverage the strengths of different learning strategies to propose the CL model that suffers less from catastrophic forgetting [3], [100], [101].

## III. CONTINUAL LEARNING FORMULATION

### A. CL-ROBUST FEATURES

In this paper, we focus on a rehearsal-based CL strategy and distill CL-robust features to construct a robust rehearsal memory, eliminating the need for reliance on an Oracle CL model [39]. Following [48], we define a time-varying *feature* to be a function mapping from the input space $\mathcal{X}_t$ at the time $t$ to the real numbers, with the set of all features, thus being $\mathcal{F}_t = \{f_t : \mathcal{X}_t \rightarrow \mathbb{R}\}$. Input-label pairs $(x_t, y_t) \in \mathcal{X}_t \times \{\pm 1\}$ are sampled from a data distribution $D_t$ at time or task $t$. For convenience, we assume that the features in $\mathcal{F}_t$ are zero-mean and unit-variance, i.e., $E_{(x_t,y_t) \sim D_t}[f_t(x_t)] = 0$ and $E_{(x_t,y_t) \sim D_t}[f_t(x_t)^2] = 1$, for all $t$.

#### 1) $\rho$-USEFUL FEATURES

A feature at a given time or task $t$ is called $\rho_t$-useful (where $\rho_t > 0$) if it correlates with the true label in expectation, that is,

$$\mathbb{E}_{(x_t,y_t) \sim \mathcal{D}_t}[y_t f_t(x_t)] \geq \rho_t, \quad (1)$$

#### 2) $\gamma$-ROBUST FEATURES

Suppose we have a feature $f_t$ that is $\rho_t$-useful at task $t$. We refer to feature $f_t$ as a *CL-robust* feature (formally a $\gamma_t$-robustly useful feature for $\gamma_t > 0$) if, under continual learning conditions, $f_t$ remains a $\gamma_t$-robustly useful for tasks $(t - \tau)$, with $0 \leq \tau < t$. Formally, if we have that

$$\mathbb{E}_{(x_{t-\tau}, y_{t-\tau}) \sim \mathcal{D}_{t-\tau}}[y_{t-\tau} f_t(x_{t-\tau})] \geq \gamma_t, \forall \tau = 0, \cdots, t-1, \quad (2)$$

where $\forall$ denotes "for all". Equation (2) states that the CL model at time $t$ retains knowledge of previous tasks. The next section provides a formal definition for distilling CL-robust features.

### B. CL-ROBUST (CLR) FEATURE DISTILLATION

The fundamental premise of our proposed framework posits the existence of CL-robust features as useful signals for CL without forgetting. We provide empirical evidence by distilling the CL-robust features to substantiate this hypothesis. Within our formal framework and a given CL model at task $t$, our objective is to construct CL-robust samples for each class $c$ at task $t$, denoted as $x_{\text{clr},t}^{[c]}$, by solving the below optimization problem. To describe this distillation process, let $f_{\theta_t}$, represent the CL model at task $t$. Let $x_t^{[c]}$ denote a sample of the $c^{th}$ class for which the corresponding

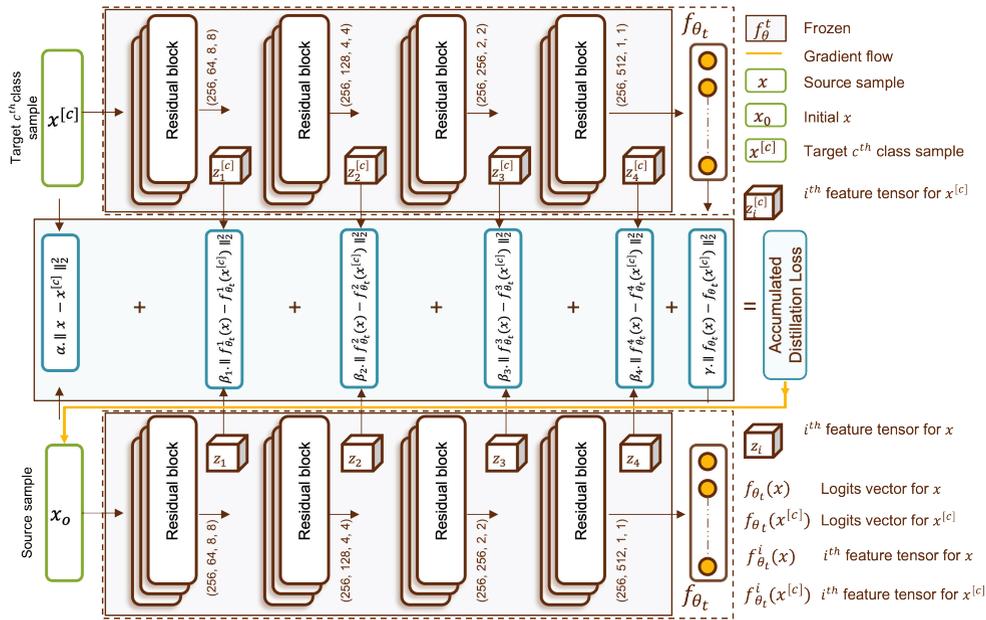

**FIGURE 1.** Diagram illustrating the CLR feature distillation. Source $x$ and target $x^{[c]}$ samples are fed through the trained CL model, $f_{\theta_t}$, at task $t$, producing intermediate feature tensors $f^i_{\theta_t}(x)$ and $f^i_{\theta_t}(x^{[c]})$ after each $i^{th}$ residual block of ResNet, respectively. The distillation loss is calculated in the input, feature, and prediction spaces. The $L_2$ norm is calculated between source $x$ and target $x^{[c]}$ in the input space. The $L_2$ norm among the $i^{th}$ corresponding feature tensors of $x$ and $x^{[c]}$, i.e., $f^i_{\theta_t}(x)$ and $f^i_{\theta_t}(x^{[c]})$, are calculated. The $L_2$ norm between the logit vectors of $x$ and $x^{[c]}$, i.e., $f_{\theta_t}(x)$ and $f_{\theta_t}(x^{[c]})$, is computed in the prediction space. The computed losses in the input, feature, and prediction spaces define the total accumulated loss. The accumulated loss is minimized with respect to $x$, i.e., the gradients are back-propagated only to $x$. Through the optimization, the $x$ is robustified into $x^{[c]}_{clr}$, which not only encapsulates the knowledge of $f_{\theta_t}$ across the input, feature, and prediction spaces, into the CL-robust sample, $x^{[c]}_{clr}$ (i.e., $x \xrightarrow{\text{robustified}} x^{[c]}_{clr}$) but it also has similar features to $x^{[c]}$.

CL-robust sample is being constructed using $f_{\theta_t}$. We refer to $x^{[c]}_t$ as the *target* sample. We construct the CL-robust sample as follows.

$$x^{[c]}_{clr,t} = \underset{x}{\arg\min}\, \mathcal{L}(x; x^{[c]}_t, f_{\theta_t}), \quad (3)$$

where $\mathcal{L}$ is the *distillation loss* to be defined in the sequel. Let $x_0$ be the initial starting point in optimization (3), referred to as the *source* sample. $x_0$ is randomly sampled from the current dataset with a class label distinct from the target class $x^{[c]}_t$. Intuitively, the distillation process (continuously) moves the source sample to the target sample with the objective of acquiring a robust sample. This robust sample exhibits proximity to the target sample in the feature space, sharing the same class while retaining a seed from the source sample, which may emanate from a historical dataset in the memory - specifically, a preceding dataset. Consequently, This distillation process produces a somewhat surreal sample characterized by a seed rooted in the past yet adeptly preserving the salient features of the current target sample. This engendered robust sample will subsequently be conveyed to the next stage in the memory pipeline, denoted as the robust memory.

We provide further details on the distillation loss $\mathcal{L}$, as outlined in Eq. (3), which comprises of three terms, $\mathcal{L}_i$, $\mathcal{L}_f$, and $\mathcal{L}_p$ i.e., $\mathcal{L} = \mathcal{L}_i + \mathcal{L}_f + \mathcal{L}_p$. These terms represent the distillation losses in the input, feature, and prediction spaces, respectively. The initial term, denoted as $L_i$, utilizes the $L_2$ norm to ensure proximity of the CL-robust sample to the target $x^{[c]}_t$ in the input space, adhering to the Euclidean distance metric.

$$\mathcal{L}_i(x; x^{[c]}_t) = \alpha \|x - x^{[c]}_t\|^2_2, \quad (4)$$

where $\alpha > 0$ represents the strength of the regularization. The second term, denoted as $L_f$, characterizes the distillation process within the feature space by ensuring the proximity of all features of the CL-robust sample to the features of the target sample, measured in the $L_2$ norm sense, i.e.,

$$\mathcal{L}_f(x; x^{[c]}_t) = \sum_{i=1}^{n} \beta_i \|f^i_{\theta_t}(x) - f^i_{\theta_t}(x^{[c]}_t)\|^2_2, \quad (5)$$

where $n$ is the number of the feature tensors of the CL model $f_{\theta_t}$, $f^i_{\theta_t}$ is the $i^{th}$ feature tensor of $f_{\theta_t}$, and $\beta_i > 0$ is a scaling factor for feature tensor $i$. The last term, $L_p$, in the distillation loss $\mathcal{L}$, performs the distillation within the prediction space as follows:

$$\mathcal{L}_p(x; x^{[c]}_t) = \gamma \|f_{\theta_t}(x) - f_{\theta_t}(x^{[c]}_t)\|^2_2, \quad (6)$$

where $f_{\theta_t}(x)$ represents the logit of sample $x$, and $\gamma > 0$ is a scaling factor. Figure 1 depicts a diagram detailing all three types of distillation losses. Note that the model $f_{\theta_t}$ is not a one-to-one mapping; the distillation loss identifies a sample that transforms the source $x_0$ into a sample within the vicinity of $x_t^{[c]}$, differing from it but possessing analogous features and logit (or class). Figure 2 illustrates the feature distillation process for the airplane class over 30, 000 epochs, while Algorithm 1 presents the pseudocodes outlining the CL-robust feature distillation process.

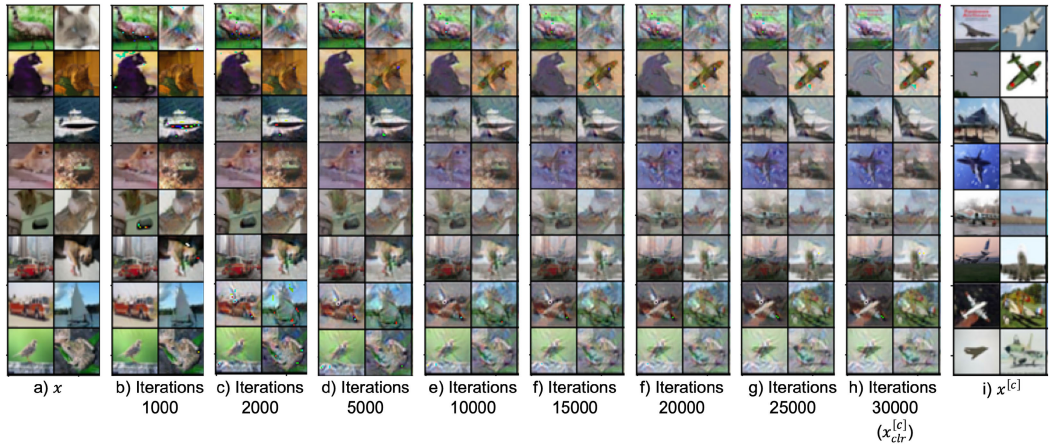

**FIGURE 2.** Sub-figure (a) and (i) display the source $x$ and target $x^{[c]}$ samples for the $c^{th}$ class, i.e., c=airplane, respectively. Sub-figures (b) to (h) illustrate the visual representation of the distillation process for the CL-robust samples pertaining to the airplane class over 30,000 epochs. Sub-figure (h) presents the final CL-robust samples for the airplane class, which are characterized by dream-like minimal representations of the airplane class.

CL-robust feature distillation occurs post-training, aiming to distill resilient features tailored to each class. These features are intended for utilization in memory during the subsequent task-learning phase. Figure 4 depicts the schematic of the proposed method, illustrating that robust feature distillation is performed after the completion of each task and prior to the learning of the next task. Algorithm 1 present the pseudocode for distilling the CL-robust samples.

### C. CL-ROBUST FEATURE RE-CONSOLIDATION

Recall that we are using a rehearsal-based CL approach. We perform CL-robust feature distillation (post-training) for the current samples $x_t$. The memory of the model consistently comprises original samples and their corresponding CL-robust samples. On the other hand, feature re-consolidation is performed for the samples within the memory. Feature re-consolidation is basically feature distillation of the samples in the memory where, for each target sample $x_t^{[c]}$, the source $x_0$ is chosen to be CL-robust sample memory $x_{\text{clr},t-1}^{[c]}$, i.e., $x_{\text{clr},t-1}^{[c]} \xrightarrow{\text{re-consolidated}} x_{\text{clr},t}^{[c]}$. Figure 3 depicts the re-consolidated CL-robust samples for the airplane class, while Algorithm 1 outlines the pseudocodes for CL-robust feature re-consolidation. Moreover, Figures 8, 9, and 10 present CL-robust samples characterized by dream-like abstract features [49], [56], [57], [58]. Notably, the CL-robust samples display characteristics akin to human dream imagery. These

**Algorithm 1** Feature Distillation

**Require:** $f_{\theta_t}, x, x_t^{[c]}, \alpha, \beta, \gamma, \eta$
    ▷ Inputs:
  $f_{\theta_t}$: trained CL model parameterized by $\theta$ at task $t$.
  $x$: source sample (initial, i.e., $x_o$)
  $x_t^{[c]}$: target sample of $c^{th}$ class
  $\alpha, \beta, \gamma$: regularization coefficients.
  $\eta$: learning rate for stochastic gradient descent (SGD).
**Ensure:** $x_{\text{clr},t}^{[c]}$   ▷ Output: CL-robust sample
1: **for** step $s = 1$ to $S$ **do**   ▷ $S$ optimization steps
2:   $\mathcal{L}_i \leftarrow \alpha \|x - x_t^{[c]}\|_2^2$   ▷ Input space
    Feed the $x$ and $x_t^{[c]}$ through $f_{\theta_t}$ to obtain their respective features tensors and logits.
3:   $\mathcal{L}_f \leftarrow \sum_{i=1}^n \beta_i \|f_{\theta_t}^i(x) - f_{\theta_t}^i(x_t^{[c]})\|_2^2$   ▷ Feature space
4:   $\mathcal{L}_p \leftarrow \gamma \|f_{\theta_t}(x) - f_{\theta_t}(x_t^{[c]})\|_2^2$   ▷ Prediction space
5:   $\mathcal{L} \leftarrow \mathcal{L}_i + \mathcal{L}_f + \mathcal{L}_p$   ▷ Accumulated loss
6:   $x \leftarrow \text{SGD}(x; f_{\theta_t}, \eta)$   ▷ SGD step over $x$ until $s \geq S$
    (i.e., $x = x + \eta \Delta_x \mathcal{L}$)
7: **end for**
8: $x_{\text{clr},t}^{[c]} \leftarrow x$   ▷ $x \xrightarrow{\text{robustified}} x_{\text{clr},t}^{[c]}$

observations parallel the phenomenon wherein the biological brain recapitulates waking experiences in a distilled form and at an accelerated timescale for memory consolidation [49], [56], [57], [58]. These neuroscientific insights endorse the rehearsal of the CL-robust samples, which facilitates the CL model's retention of knowledge from previously learned classes through the recollection and rehearsal of class abstractions with minimal details.

### D. A STEP-BY-STEP BREAKDOWN OF THE PROPOSED APPROACH

For enhanced understanding, consider a rehearsal-based CL model that learns 9 tasks sequentially, where the initial task

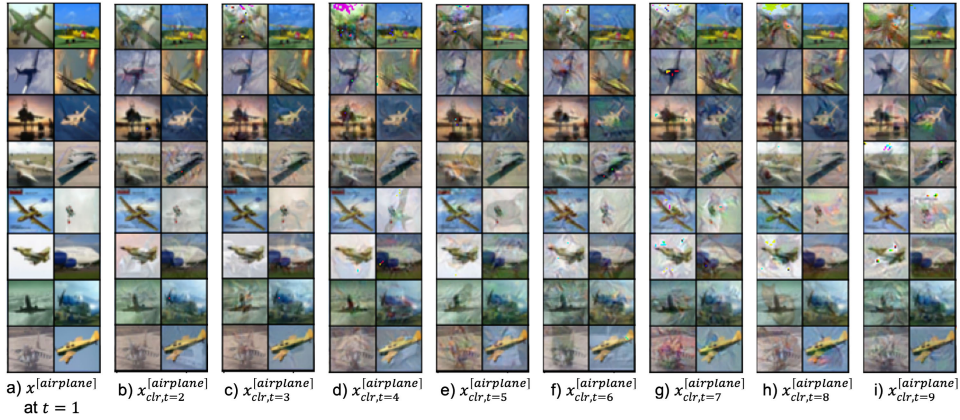

a) $x^{[airplane]}$ at $t = 1$ b) $x^{[airplane]}_{clr,t=2}$ c) $x^{[airplane]}_{clr,t=3}$ d) $x^{[airplane]}_{clr,t=4}$ e) $x^{[airplane]}_{clr,t=5}$ f) $x^{[airplane]}_{clr,t=6}$ g) $x^{[airplane]}_{clr,t=7}$ h) $x^{[airplane]}_{clr,t=8}$ i) $x^{[airplane]}_{clr,t=9}$

**FIGURE 3.** Task-wise, CL-robust memory samples are presented for airplane class. Sub-figure (a) presents the original airplane images presented to the CL model during the first tasks at time $t = 1$. Sub-figures (b) to (i) depict the CL-robust samples of the airplane class distilled after completing each subsequent task in the nine tasks configuration.

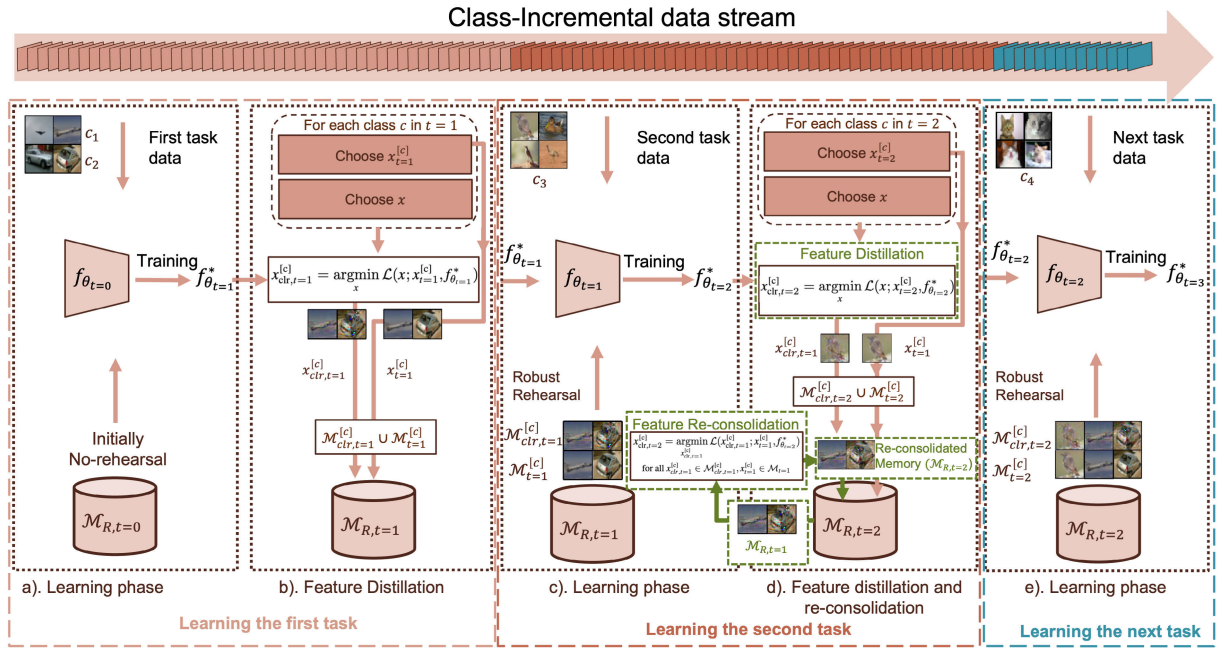

**FIGURE 4.** Schematic diagram of the proposed CL-robust rehearsal. (a) The model starts learning the first task, which encompasses two classes (airplane and car). The robust rehearsal memory, $\mathcal{M}_{R,t=0}$, is initially empty at $t = 0$. The initial model, $f_{\theta_{t=0}}$ at time $t = 0$, is trained to obtain an optimal model, $f^*_{\theta_{t=1}}$, for the first task at $t = 1$. (b) For each class in task $t = 1$, $x^{[c]}_{t=1}$ samples are chosen, and $x$ samples are randomly drawn from $\mathcal{D}_{t=1}$ with class label distinct than $c$. Next, the feature distillation is conducted using $f^*_{\theta_{t=1}}$ to construct the CL-robust samples and stored in $\mathcal{M}_{clr,t=1}$. Collectively, $\mathcal{M}^{[c]}_{clr,t=1}$ and $\mathcal{M}^{[c]}_{t=1}$ for each class at task $t = 1$, form the CL-robust memory $\mathcal{M}^{[c]}_{R,t=1}$. (c) During the learning of the second task, samples from the robust rehearsal memory, $\mathcal{M}_{R,t=1}$, are interleaved with the mini-batches of the second task. After learning the second task at the $t = 2$, the CL model $f^*_{\theta_{t=1}}$ becomes $f^*_{\theta_{t=2}}$ at the task $t = 2$. (d) Similarly, the steps discussed in (a) are utilized to perform feature distillation for each class in the second task at $t = 2$ using the $f^*_{\theta_{t=2}}$, while feature re-consolidation is performed for each previously learned class in memory. The feature re-consolidation of the CL-robust memory samples results in the updating the CL-robust memory samples from time $t = 1$, $x^{[c]}_{clr,t=1}$, into $x^{[c]}_{clr,t=2}$, i.e., $x^{[c]}_{clr,t=1} \xrightarrow{re-consolidated} x^{[c]}_{clr,t=2}$ for each previous class. The newly re-consolidated CL-robust samples, $x^{[c]}_{clr,t=2}$, supersede the previous samples, $x^{[c]}_{clr,t=1}$ for each $c^{th}$, in the $\mathcal{M}_{R,t=2}$ and are rehearsed during the learning of the subsequent task. (e) As the CL model engages in learning the subsequent task, samples from the robust rehearsal memory are interleaved with those of the subsequent task. Following the completion of subsequent tasks, feature distillation and re-consolidation are similarly performed.

involves two classes and each subsequent task introduces one additional class in a class incremental learning setting.

Figure 4 presents the schematic diagram of the proposed example.

**Algorithm 2** Robust Rehearsal
**Require:** $\mathcal{D}_t, \mathcal{M}_{R,t}, f_{\theta_{t-1}}, \alpha$
▷ Inputs:
  $\mathcal{D}_t$: dataset for task $t$
  $\mathcal{M}_{R,t}$: CL-robust memory at task $t$
  $f_{\theta_{t-1}}$: CL model parameterized by $\theta$ trained on previous task $t-1$
  $\alpha$: learning rate
**Ensure:** $f_{\theta_t}^*$  ▷ Output the trained CL model $f_{\theta_t}^*$.
1: **for** $B_t \sim \mathcal{D}_t$ **do**  ▷ $B_t$ is a mini-batch sampled from $\mathcal{D}_t$
2:   $B_R \leftarrow \text{MemoryRetrieval}(\mathcal{M}_{R,t})$ ▷ $B_R$ is from $\mathcal{M}_{R,t}$
3:   $\theta \leftarrow \text{SGD}(\theta; B_t \cup B_R, \alpha)$ ▷ SGD step over $\theta$
4: **end for**  ▷ $f_{\theta_{t-1}} \xrightarrow{\text{converged}} f_{\theta_t}^*$

### 1) LEARNING TASK 1
The initial task encompasses two classes, i.e., airplane and car, presented to a rehearsal-based CL model at task or time $t = 1$. The CL model learns the first task, resulting in $f_{\theta_{t=1}}^*$, representing the trained CL model for the initial task. The robust rehearsal memory, $\mathcal{M}_R$, at $t = 0$, is empty.

### 2) FEATURE DISTILLATION OF TASK 1
Before proceeding to the next task, choose a subset of samples for each class in the current task and perform feature distillation to generate their respective CL-robust samples. The aggregated set of distilled CL-robust samples, collectively represented as $x_{\text{clr},t=1}$, will constitute the CL-robust memory $\mathcal{M}_{\text{clr},t=1}$. The robust rehearsal memory $\mathcal{M}_{R,t=1}$ is the combination of $\mathcal{M}_{\text{clr}}$ and the classical memory, which comprises the original samples, i.e., $\mathcal{M}_R = \mathcal{M} \cup \mathcal{M}_{\text{clr}}$. Algorithm 1 provides the pseudocode of feature distillation.

### 3) LEARNING TASK 2
As in a rehearsal-based CL approach, the model advances to learn the second task using the current dataset at $t = 2$, $\mathcal{D}_2$, along with the robust memory $\mathcal{M}_{R,t=1}$ produced at $t = 1$. In this example, $\mathcal{D}_2$ contains samples from the bird class. After learning the second task, the CL model parameters update from $f_{\theta_{t=1}}^*$ to $f_{\theta_{t=2}}^*$.

### 4) FEATURE DISTILLATION OF TASK 2
At the end of training task 2, we select a subset of bird samples and perform feature distillation to obtain CL-robust bird samples. Before defining the robust memory at $t = 2$, we need to perform a re-consolidation of the car and airplane memory samples.

### 5) FEATURE RE-CONSOLIDATION AT TASK 2
Feature re-consolidation is performed for the memory samples, i.e., cars and airplanes, by distilling each sample such that each target starts with a seed equal to its corresponding CL-robust sample. Finally, the robust memory at $t = 2$ consists of original birds, distilled birds, original cars and airplanes, and re-consolidated cars and airplanes, i.e.,

$\mathcal{M}_{R,t=2} = \mathcal{M}_{t=2}^{\text{bird}} \cup \mathcal{M}_{\text{clr},t=2}^{\text{bird}} \cup \mathcal{M}_{t=1}^{\text{car}} \cup \mathcal{M}_{\text{clr},t=2}^{\text{car}} \cup \mathcal{M}_{t=1}^{\text{airplane}} \cup \mathcal{M}_{\text{clr},t=2}^{\text{airplane}}$. Algorithm 1 outlines the steps involved in the re-consolidation process. Figures 6 and 3 display the CL-robust samples for the airplane class throughout the nine learning tasks.

### 6) LEARNING THE NEXT TASKS SEQUENTIALLY
After learning each task, feature distillation of current samples, followed by memory re-consolidation, is conducted. Figure 4 provides a schematic diagram of the proposed framework, which delineates the feature distillation and re-consolidation performed after learning each task.

## IV. EXPERIMENTAL RESULTS
This section describes the experimental setup, benchmark datasets employed, the CL protocols implemented, the baseline approaches utilized, the training details, the evaluation methodology, results and discussion, and concludes by providing the limitations of the proposed robust rehearsal. Our observations indicate that baseline approaches augmented with robust rehearsal outperformed their counterparts not employing robust rehearsal on three benchmark datasets.

### A. EXPERIMENTAL SETUP
#### 1) DATASETS
We employed benchmark datasets, split CIFAR10 and split CIFAR100, in conjunction with a real-world split Helicopter Attitude dataset provided by the Federal Aviation Administration (FAA) [1], [38], [102], [102], [103], [104]. CIFAR10 is a complex dataset comprising ten distinct classes, each encompassing 6,000 images, of which 5000 are used as the training set while the remaining 1,000 are used for the test set [102]. CIFAR100, on the other hand, encompasses 100 classes, each consisting of 600 images, of which 500 were reserved for training, while the remaining 100 were utilized for testing [102]. The Helicopter Attitude dataset, characterized as a highly complex real-world dataset [103], [104], consists of video frames sourced from the camera mounted within the cockpit of a Sikorsky S-76 helicopter. These frames capture the external view through the windshield of the S-76 during ten different flights, accumulating a total flight duration of approximately seven hours. The primary objective is incremental learning of helicopter attitude prediction, utilizing the pitch and roll angle values. The ground truth values of pitch and roll angles for each frame were obtained from the helicopter's in-flight data recorder. To adapt the attitude prediction task into the classification framework, a predefined threshold ($\alpha$) was employed on the pitch and roll values to define nine different bins representing mutually exclusive nine discrete classes of helicopter attitude [103], [104]. Table 1 provides the individual attitude classes, while Figure 5 illustrates a representative sample of each attitude class. Each of the nine attitude classes comprised 25,000 images, of which 20,000 images per class were dedicated to the training and 2,500 for validation, while the remaining 2,500 images per class were designated for the testing set. Table 2 summarizes the characteristics of all three datasets.

TABLE 1. A threshold value, denoted by $\alpha = 3$, is applied to Flight Data Recorder (FDR) metrics for pitch and roll, establishing nine mutually exclusive discrete attitude classes. Abbreviations used: NU - nose up, ND - nose down, RP - roll positive; RN - roll negative, and L - level or steady-state.

| Class | Description | Pitch (P) | Roll (R) |
|---|---|---|---|
| 0 | NU | $P > \alpha$ | $-\alpha \leq R \leq +\alpha$ |
| 1 | ND | $P < -\alpha$ | $-\alpha \leq R \leq +\alpha$ |
| 2 | RP | $-\alpha \leq P \leq +\alpha$ | $R > \alpha$ |
| 3 | RN | $-\alpha \leq P \leq +\alpha$ | $R < -\alpha$ |
| 4 | NU & RP | $P > \alpha$ | $R > \alpha$ |
| 5 | NU & RN | $P > \alpha$ | $R < -\alpha$ |
| 6 | ND & RP | $P < -\alpha$ | $R > \alpha$ |
| 7 | ND & RN | $P < -\alpha$ | $R < -\alpha$ |
| 8 | L | $-\alpha \leq P \leq +\alpha$ | $-\alpha \leq R \leq +\alpha$ |

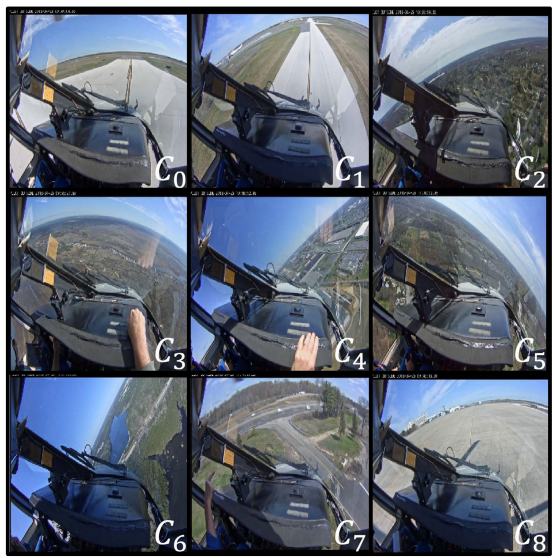

FIGURE 5. Helicopter attitude dataset: Sample images for nine different classes. Class labels ($C_i$), as defined in Table 1, are given in the lower right corner of each image.

TABLE 2. Datasets statistics.

| Dataset | CIFAR10 | CIFAR100 | Helicopter Attitude |
|---|---|---|---|
| Input size | $3 \times 32 \times 32$ | $3 \times 32 \times 32$ | $3 \times 128 \times 128$ |
| # Classes | 10 | 100 | 9 |
| # Training samples per class | 4500 | 450 | 20,000 |
| # Validation samples per class | 500 | 50 | 2,500 |
| # Testing samples per class | 1000 | 100 | 2,500 |

2) PROTOCOLS
We followed the well-established class-incremental learning (CIL) framework to partition the datasets into multiple tasks or steps [1], [105]. The CIL is considered one of the most challenging protocols in CL, as it effectively emulates real-world scenarios, where the model learns sequentially from streaming tasks without prior knowledge of task identification [105], [106]. Following the CIL protocol, the CIFAR10 was divided into nine, five, and two tasks or steps, while the Helicopter Attitude dataset was divided into eight, five, each with a fixed memory size of 1024 exemplars [78], [103]. For CIFAR100, we utilized two well-established CIL protocols: (1) B0 (base 0) [19], where the model is trained over 5, 10, and 20 steps, introducing 20, 10, and 5 new classes at each step, respectively, while maintaining a fixed memory size of 2,000 exemplars; and (2) B50 (base 50) [107], in which the model is initially trained on 50 classes, followed by sequential learning of the remaining 50 classes, introduced in increments of 5 and 10 classes per step, while the memory size remains fixed at 20 exemplars per class.

3) BASELINE MODELS
Initially, we assessed the performance of the robust rehearsal using the same baseline approaches as outlined in [39], which included: 1) A standard replay-based strategy, which stores and rehearses the subset of data from previous tasks during the learning of subsequent tasks [20]; 2) Dynamically Expanding Representation (DER) that increases the network capacity to accommodate new tasks while preserving previously acquired representations [30]; 3) Pooled Outputs Distillation (PODNet), applying spatial distillation loss to maintain consistent representations across sequential tasks, thus mitigating substantial shifts in feature representation [29]; 4) A simple fine-tuning without integrating specific techniques for retaining knowledge of the previous tasks, serving as the lower bound performance benchmark for CL models; and 5) a multi-task or joint learning model, which concurrently learns all tasks to attain the highest performance and serve as the upper-bound performance benchmark. The proposed robust rehearsal is based on a rehearsal-based strategy and is compatible with a range of CL approaches. To demonstrate its practical versatility, we broadened our baselines to incorporate additional CL approaches, particularly for CIFAR100 and the Helicopter Attitude dataset. The additional approaches are as follows: 1) Incremental Classifier and Representation Learning (iCaRL) retaining class-specific sample subsets for centroid approximation and employing nearest class mean classifiers [19]; 2) Weight Aligning (WA) using knowledge distillation for old class discrimination and bias adjustment in the fully connected layer without extra parameters [108]; 3) Feature boOSTing and comprEssion for class-incRemental learning (FOSTER) [109], combining dynamic architecture with distillation, similar to the DER [30], for distilling both the preceding and current networks into a single, consolidated model; and lastly, 4) Bias Correction (BiC) [83] utilizing a bias correction layer in the fully connected layer to effectively address biases and imbalances between old and new classes through a linear model.

### 4) TRAINING DETAILS

We utilized the PyTorch framework [110] in conjunction with the publicly available PyCIL codebase [111] for our implementation. We employed standard residual network architectures [112] with Xavier initialization [113]. For the CIFAR-10 and CIFAR-100 datasets, we opted for ResNet-34, while for the Helicopter Attitude dataset, we utilized ResNet-50 architectures [112]. Across all experiments, a consistent batch size of 256 was employed. All models were trained using a stochastic gradient descent optimizer with an initial learning rate of 0.01, a momentum of 0.9, and a learning rate decay of 0.1. For classification loss, the standard cross-entropy loss function is employed. The training epoch for the initial task was set to 200 epochs, whereas it was set to 128 for the following tasks. The standard data augmentation, such as random cropping, brightness, and contrast adjustment, was applied to the current task data and original memory samples in the rehearsal memory [114], [115], while the CL-robust samples residing in the rehearsal memory were interleaved without undergoing data augmentation. Feature distillation and re-consolidation were performed employing a stochastic gradient descent optimizer, initialized with a learning rate of 0.001 and momentum of 0.9, for over 30,000 iterations. Cosine annealing was used to reduce the learning rate throughout the feature distillation and re-consolidation using Algorithm 1. The other hyper-parameters specific to each baseline were set to their default configurations, as specified in their respective papers [19], [29], [30], [83], [108], [109], and implemented in PyCIL [111] to ensure the optimal performance of each baseline and to derive maximal benefit from the integration of the robust rehearsal memory.

### 5) EVALUATION

We employ the widely used metric of average classification accuracy (ACA) to quantify the performance of the CL models [1], [19], [21], [23], [27], [78], which is calculated by evaluating the final trained model on all the learned tasks. Formally, the average classification accuracy can be defined as:

$$ACA = \frac{1}{T}\sum_{i=1}^{T} R_{T,i}, \qquad (7)$$

where $R$ represents the accuracy, $T$ is the total number of tasks, and $i$ represents the task index.

### B. RESULTS AND DISCUSSION

Table 3 provides a comparative analysis of the performance of baseline approaches utilizing robust rehearsal and their counterparts not employing robust rehearsal across nine, five, and two tasks on the CIFAR10 dataset. The approaches are categorized into three categories: 1) the standard baseline approaches, denoted without any suffix, were trained using their respective proposed methodologies [19], [29], [30], [83], [108], [109]; 2) the baseline approaches, denoted with the CLR suffix, trained on the pre-distilled robustified CIFAR10 [39]; and 3) robust rehearsal approaches, denoted with the RR suffix, were trained utilizing the proposed robust rehearsal strategy. As can be seen in Table 3, approaches employing robust rehearsal surpassed the performance of the standard baseline approaches, denoted without a suffix, across nine, five, and two task configurations. Similarly, the approaches based on robust rehearsal also exceeded the performance of the CLR counterparts, which were trained on a pre-distilled (using pre-trained Oracle model) CL-robustified CIFAR10, as elaborated in [39]. The suboptimal performance of the CLR approaches underscores the limitation in the transferability of pre-distilled features in the context of sequential learning. On the other hand, the robust rehearsal approach rectified this limitation inherent to the CLR approaches and accommodated the time-varying nature of the CL-robust features through feature re-consolidation [39]. Moreover, as shown in Table 3, the robust rehearsal-based (i.e., Replay-RR) leveraged feature distillation and re-consolidation to a greater extent compared to approaches based on dynamic architecture (e.g., DER-RR) or spatial distillation approaches (e.g., PODNet-RR). A plausible explanation for this observation is that freezing the network architecture or the imposition of additional regularizing terms may hinder the CL model's capacity to adapt to the time-varying CL-robust feature as the model sequentially learns new tasks. In contrast, with robust rehearsal, the re-distillation for CL-robust samples accommodated for such time-varying features. Subsequently, rehearsing the re-consolidated CL-robust samples reinforced the CL model in maintaining focus on the CL-robust features of previously learned classes, thereby preventing catastrophic forgetting. Figures 6 and 3 present the CL-robust features. We deduced that the imposition of any form of regularization, either in the form of distillation or penalizing the model

**TABLE 3.** The comparative performance of CL approaches on the CIFAR10 dataset. The approaches are categorized into three categories: (1) standard baselines (no suffix), representing methods trained using their respective proposed strategies; (2) baselines with suffix CLR, indicating approaches trained on the pre-distilled robustified version of the dataset as per [39]; and (3) approaches with RR suffix, utilizing the proposed robust rehearsal strategy. The robust rehearsal strategy, which involves rehearsing the CL-robust sample during each task, consistently improved accuracy across all three class incremental learning steps (9, 5, and 2 steps) on the CIFAR10 dataset.

| CL Method | Split-CIFAR10 | | |
|---|---|---|---|
| | 9 steps | 5 steps | 2 steps |
| Joint | | 94.8±0.61 | |
| Fine-tune | 11.11±0.68 | 18.54±0.71 | 45.31±1.6 |
| Replay [20] | 55.06 ±1.90 | 62.18 ±1.30 | 63.9 ±0.97 |
| Replay-CLR [39] | 75.86 ±1.20 | 78.29 ±0.93 | 80.77 ±1.09 |
| Replay-RR | **77.34** ±1.43 | **80.12** ±1.39 | **84.11** ±1.66 |
| DER [30] | 52.42 ±1.50 | 65.65 ±0.72 | 71.60 ±0.94 |
| DER-CLR [39] | 70.13 ±1.10 | 81.74 ±0.65 | 87.32 ±0.90 |
| DER-RR | **72.39** ±1.78 | 74.16 ±0.97 | 83.92 ±0.84 |
| PODNet [29] | 55.14 ±1.20 | 59.62 ±0.74 | 80.09 ±1.16 |
| PODNet-CLR [39] | 72.71 ±1.52 | 76.94 ±0.63 | **85.82** ±1.23 |
| PODNet-RR | **74.23** ±0.76 | **77.88** ±0.89 | 84.54 ±1.25 |

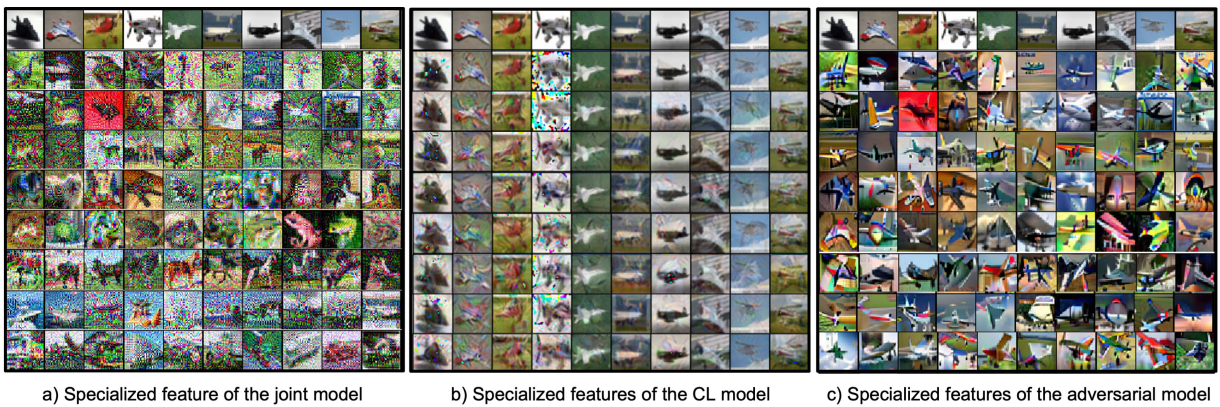

a) Specialized feature of the joint model     b) Specialized features of the CL model     c) Specialized features of the adversarial model

**FIGURE 6.** Comparative visualization of the features of three models for the airplane class from the CIFAR10 dataset, trained via three distinct training protocols: joint or multi-task learning $f_{\theta_{MT}}$, CL model $f_{\theta_t}$, and adversarial training model $f_{\theta_{AT}}$ are presented in sub-figures a), b) and c) respectively. For the continual learning model in sub-figure b), the first row presents airplane samples as they appear in the initial task. The subsequent rows illustrate memory CL-robust samples for the airplane class obtained through feature re-consolidation as the CL model learns each subsequent task. Conversely, in both joint and adversarial robust training protocols, the models learn the dataset in one step (i.e., without sequential learning). Therefore, a random batch of images is utilized to conduct feature distillation for both joint and adversarial robust models. Notably, the models trained under different training protocols learned distinct features for the same class. This highlights that the training protocol and optimization objective highly guide learning the diverse features. Notably, the CL model, $f_{\theta_t}$, features demonstrate the time-varying nature or change over the course of sequential learning.

**TABLE 4.** The comparative performance of the CL approaches on the CIFAR100 dataset was evaluated using B0 and B50 protocols. Under B0 protocol, the model undergoes training in 5 steps (with 20 new classes per step), 10 steps (with 10 new classes per step), and 20 steps (with five new classes per step) with a consistent memory size of 2,000 exemplars. Conversely, the B50 involves initial training on 50 classes, followed by continual learning on the remaining 50 classes in increments of 5 and 10 classes per step, maintaining a memory size of 20 exemplars per class. The approaches are divided into (1) Standard Baselines (no suffix), using their respective proposed strategies, and (2) Approaches with RR suffix, employing the robust rehearsal strategy. The approaches augmented with CL-robust sample rehearsal consistently improved the performance of the corresponding baseline models trained without the integration of robust rehearsal.

| CL Method | Split-CIFAR100 B0 | | | Split-CIFAR100 B50 | |
|---|---|---|---|---|---|
| | 5 steps | 10 steps | 20 steps | 5 steps | 10 steps |
| Joint | | | 84.45 ±2.36 | | |
| Fine-tune | 27.8 ±2.29 | 22.4 ±1.91 | 19.5 ±2.13 | 13.49 ±1.98 | 11.34 ±2.71 |
| iCaRL [19] | 68.12 ±2.25 | 66.74 ±2.79 | 64.01 ±1.92 | 59.29 ±2.32 | 50.82 ±1.82 |
| iCaRL-RR | **72.87** ±1.34 | **73.66** ±1.24 | **70.91** ±1.53 | **66.24** ±1.09 | **59.31** ±1.37 |
| PodNET [29] | 69.54 ±1.01 | 62.22 ±1.27 | 57.65 ±1.87 | 67.39 ±2.74 | 66.23 ±1.59 |
| PodNET-RR | **77.51** ±1.56 | **72.04** ±1.75 | **66.65** ±2.22 | **75.87** ±2.05 | **74.89** ±1.85 |
| DER [30] | 72.56 ±2.68 | 71.44 ±1.89 | 70.22 ±2.34 | 68.61 ±2.11 | 69.02 ±1.88 |
| DER-RR | **76.39** ±2.25 | **79.09** ±2.31 | **81.74** ±1.96 | **77.63** ±1.94 | **82.89** ±2.15 |
| WA [108] | 70.23 ±1.81 | 67.34 ±2.22 | 63.77 ±2.72 | 65.33 ±1.82 | 60.11 ±1.84 |
| WA-RR | **74.65** ±1.93 | **72.25** ±1.95 | **68.56** ±1.87 | **73.92** ±1.74 | **67.17** ±2.15 |
| FOSTER [109] | 70.97 ±1.83 | 69.19 ±2.01 | 64.37 ±2.51 | 68.58 ±1.91 | 67.17 ±1.71 |
| FOSTER-RR | **75.10** ±1.66 | **74.09** ±1.85 | **71.39** ±2.47 | **74.61** ±1.78 | **71.49** ±1.91 |
| BiC [83] | 67.35 ±1.75 | 65.78 ±1.99 | 62.41 ±2.14 | 61.20 ±2.21 | 55.98 ±2.84 |
| BiC-RR | **70.51** ±1.81 | **72.01** ±2.11 | **68.63** ±1.79 | **67.91** ±1.93 | **64.71** ±2.30 |

parameters, which attempts to preserve the old representation or closely align them with the previous solution, adversely impacts the performance of the CL model in future tasks.

Table 4 presents a comparative analysis of the performance of baseline approaches utilizing robust rehearsal and their counterparts not employing robust rehearsal on the split-CIFAR100 dataset under two protocols, B0 and B50. The standard baseline approaches, denoted without any suffix, were trained in accordance with their respective prescribed methodologies as outlined in [19], [29], [30], [83], [108], and [109]. Conversely, the baseline approaches that incorporated the proposed robust rehearsal strategy are denoted with the RR suffix. Notably, across both protocols, i.e., B0 and B50, all six baseline approaches that employ the robust rehearsal strategy consistently demonstrate superior performance compared to the standard baseline approaches. Moreover, it was noted that robust rehearsal demonstrates comparatively greater effectiveness in larger task sequences, such as those with 20 splits, as compared to shorter sequences, i.e., 5 splits. Additionally, pre-training the base model on 50 classes using the Base 50 protocol, followed by subsequent learning in 5-splits and 10-splits, resulted in a noticeable improvement in the model's average accuracy compared to training under the B0 protocol. Such improvement can be attributed to the fact that the B50 protocol, where the CL model is pre-trained on an initial set of 50 classes, possesses more refined initial features than the B0 protocol, where the learning begins from Xavier initialization [113].

**TABLE 5.** The comparative performance of CL approaches across nine, five, and two tasks on the Helicopter Attitude dataset is illustrated. The approaches are categorized into (1) Standard Baselines, following their respective proposed strategies, and (2) Approaches with the RR suffix, utilizing the robust rehearsal strategy. The first column lists the names of the approaches, while the second, third, and fourth columns display the comparative average accuracy of the CL approaches in eight, five, and three task configurations of class incremental learning, respectively. The approaches augmented with CL-robust sample rehearsal consistently demonstrated improved performance compared to baseline models that underwent training without robust rehearsal. These enhanced performances of the robust rehearsal-based approaches underscore the effectiveness of incorporating CL-robust rehearsal as a methodological enhancement to bolster the performance of CL models.

| CL Method | S-76 Helicopter Attitude Dataset | | |
|---|---|---|---|
| | 8 steps | 5 steps | 3 steps |
| Joint | | 88.6±0.61 | |
| Fine-tune | 12.9±0.83 | 21.65±0.68 | 30.77±1.12 |
| iCaRL [19] | 44.98 ±0.73 | 52.75 ±1.15 | 61.28 ±0.93 |
| iCaRL-RR | **48.32** ±0.82 | **60.45** ±1.45 | **68.21** ±1.41 |
| PODNet [29] | 51.27 ±1.76 | 62.33 ±2.06 | 79.12 ±0.87 |
| PODNet-RR | **58.96** ±1.71 | **69.15** ±2.21 | **83.25** ±1.37 |
| DER [30] | 50.83 ±1.24 | 65.88 ±1.71 | 77.20 ±0.82 |
| DER-RR | **60.97** ±1.13 | **72.52** ±0.99 | **85.51** ±1.39 |
| WA [108] | 43.61 ±1.61 | 57.83 ±0.89 | 64.87 ±1.29 |
| WA-RR | **45.72** ±1.36 | **58.89** ±1.02 | **65.60** ±1.10 |
| FOSTER [109] | 48.69 ±1.28 | 58.77 ±1.66 | 65.73 ±0.86 |
| FOSTER-RR | **54.59** ±0.94 | **60.55** ±1.39 | **69.64** ±0.91 |
| BiC [83] | 37.51 ±0.94 | 48.30 ±1.71 | 65.23 ±0.82 |
| BiC-RR | **41.63** ±1.08 | **50.93** ±1.78 | **66.35** ±1.19 |

Table 5 presents a comparative analysis of the performance of baseline approaches utilizing robust rehearsal and their counterparts not employing robust rehearsal on the split Helicopter Attitude dataset across eight tasks, five tasks configurations [103], [104]. A consistent observation indicated that the CL baseline approaches trained with the robust rehearsal strategy outperformed their standard counterparts. In other words, the robust rehearsal strategy maximally utilizes the feature distillation and re-consolidation strategies, corroborated by its maintenance of the highest average accuracy compared to its counterparts. These findings underscore the pivotal role of combining rehearsal strategy to distill and then rehearse the CL-robust samples, thereby validating the efficacy of incorporating CL-robust rehearsal.

1) EFFECT OF THE REHEARSAL BUFFER SIZE

We assessed the impact of rehearsing the CL-robust samples within varying rehearsal memory sizes. The results of the experiments are presented in Table 6, illustrating that robust rehearsal achieved higher accuracy compared to naive rehearsal methods across ten rehearsal memory sizes.

Recall that the sole difference between robust rehearsal and naive rehearsal-based training was the addition or no addition of CL-robust samples in the rehearsal memory, respectively. An additional noteworthy observation from our analysis is that robust rehearsal achieves higher accuracy in comparatively smaller memory sizes and consistently maintains higher average accuracy over ten memory sizes. Figure 7 displays the average accuracy of all ten experiments across ten memory sizes for nine, five, and two task configurations.

2) EFFECT OF THE NUMBER OF TASKS

We conducted experiments utilizing the CIFAR10, Helicopter Attitude, and CIFAR100 to assess the effectiveness of the robust rehearsal across varying numbers of sequential tasks. Tables 3, 5, and 4 present results for robust rehearsal across nine, five, and two tasks for CIFAR10, eight, five, and three for Helicopter Attitude, and five, ten, and twenty tasks for the CIFAR100 datasets, respectively. The methods employing robust rehearsal, denoted with the suffix RR, surpassed the performance of the baseline methods that did not employ the robust rehearsal, denoted without any suffix, across a range of task numbers on CIFAR10, Helicopter Attitude, and CIFAR100 datasets. The methods augmented with robust rehearsal consistently improved the performance of the baseline without the integration of robust rehearsal. This consistent advantage of robust rehearsal across varying task numbers offers valuable insights into its scalability and adaptability, underscoring its suitability for developing CL methods capable of effectively learning a large sequence of tasks.

3) DIVERSE FEATURE SPECIALIZATION

We empirically demonstrate that deep learning models acquire diverse features when trained with distinct training protocols to solve respective optimization objectives. To this end, we first trained a joint or multi-task model, denoted as $f_{\theta_{MT}}$, following the standard joint learning protocol. Subsequently, we trained the CL oracle model, as detailed in [39], employing the robust rehearsal strategy. For the adversarially robust model, we leveraged a state-of-the-art publicly available model and publically accessible,[1] as described in [116]. Figure 6 presents the learned or specialized features of the three trained models, as obtained using Algorithm 1, which presents a novel perspective, revealing that diverse features were acquired by $f_{\theta_{MT}}$, $f_{\theta_t}$, and $f_{\theta_{AT}}$, each employing distinct optimization objectives during their training phase. For instance, the primary objective of joint or multi-task models is to maximize validation accuracy by leveraging any available features with high generalizability from the training dataset without differentiating between the robust and non-robust quality of the features, which resulted in models that achieve higher validation accuracy, albeit at the expense of robustness [48]. On the other hand, adversarially robust models prioritize a formidable defense against adversaries, necessitating a balance between training accuracy and model robustness [48]. These models predominantly learn adversarially robust features, which are resistant to adversarial attacks, and exhibit little or no reliance on non-robust features to obtain a model with enhanced adversarial robustness. Conversely, CL models, which learn sequentially, must balance plasticity and stability by learning specialized features, such as CL-robust features, to mitigate catastrophic forgetting while simultaneously facilitating the sequential acquisition of new tasks. Figure 6 displays the specialized features of models trained under three distinct training protocols: joint or multi-task learning, continual learning, and adversarial learning, and emphasizes that features optimized for a specific objective may not necessarily be suitable for other optimization goals. Furthermore, it has been shown that the available features in the dataset played a vital role in determining the model's overall robustness, generalizability, and performance [40], [48]. These insights further emphasize the importance of rehearsing CL-robust samples using robust rehearsal to mitigate catastrophic forgetting as the model undergoes sequential learning effectively. Figures 8, 9, and 10 display the CL-robust samples for various classes within the CIFAR10 dataset.

---

[1] https://github.com/RobustBench/robustbench

TABLE 6. The average accuracy for ten CL models trained using the naive rehearsal and robust rehearsal strategies on the CIFAR-10 dataset. The first column displays the names of the CL methods. The naive rehearsal employed a standard rehearsal-based strategy for training, while the robust rehearsal was trained using the proposed robust rehearsal strategy. The second column indicates the rehearsal memory size. Subsequent columns provide the average accuracy for nine, five, and two tasks in class incremental settings, respectively. Each experiment is repeated five times with different initializations to obtain meaningful estimates of the performance of experiments across varying memory sizes.

| CL Method | Memory | Split-CIFAR10 | | |
|---|---|---|---|---|
| | | Nine Tasks | Five Tasks | Two Tasks |
| Multi-Task | not applicable | | 94.8±3.11 | |
| Fine-tune | not applicable | 11.41±2.68 | 18.54±1.71 | 45.31±1.9 |
| Naive Rehearsal | 32 | 14.81±3.10 | 25.11±2.68 | 47.62±2.33 |
| Robust Rehearsal | | **20.78**±2.87 | **33.1**±3.97 | **48.69**±1.51 |
| Naive Rehearsal | 64 | 20.92±3.96 | 29.11±3.67 | 48.47±2.19 |
| Robust Rehearsal | | **29.12**±2.55 | **37.81**±2.10 | **51.59**±2.27 |
| Naive Rehearsal | 128 | 28.91±3.3 | 36.61±3.33 | 52.92±3.87 |
| Robust Rehearsal | | **42.77**±3.44 | **42.78**±4.57 | **57.86**±3.18 |
| Naive Rehearsal | 256 | 31.54±4.01 | 45.82±4.15 | 56.72±2.98 |
| Robust Rehearsal | | **47.01**±2.59 | **53.92**±2.64 | **61.76**±2.53 |
| Naive Rehearsal | 512 | 41.91±4.56 | 53.92±4.36 | 60.24±3.86 |
| Robust Rehearsal | | **59.20**±4.56 | **63.59**±3.77 | **68.13**±2.83 |
| Naive Rehearsal | 1024 | 55.77±3.92 | 59.38±2.85 | 65.48±3.91 |
| Robust Rehearsal | | **76.42**±3.23 | **76.46**±4.11 | **75.55**±3.21 |
| Naive Rehearsal | 2048 | 65.94±4.50 | 67.67±3.71 | 72.57±2.85 |
| Robust Rehearsal | | **80.56**±2.32 | **77.34**±3.91 | **79.22**±2.48 |
| Naive Rehearsal | 4096 | 73.26±3.44 | 74.21±2.94 | 76.51±3.45 |
| Robust Rehearsal | | **85.06**±3.22 | **83.53**±3.03 | **82.19**±2.59 |
| Naive Rehearsal | 8192 | 79.59±2.85 | 75.33±1.94 | 79.12±2.97 |
| Robust Rehearsal | | **87.42**±3.45 | **85.19**±2.91 | **87.21**±1.89 |
| Naive Rehearsal | 16382 | 82.48±1.53 | 80.45±2.26 | 82.22±2.84 |
| Robust Rehearsal | | **88.19**±1.85 | **88.94**±1.90 | **89.04**±1.97 |

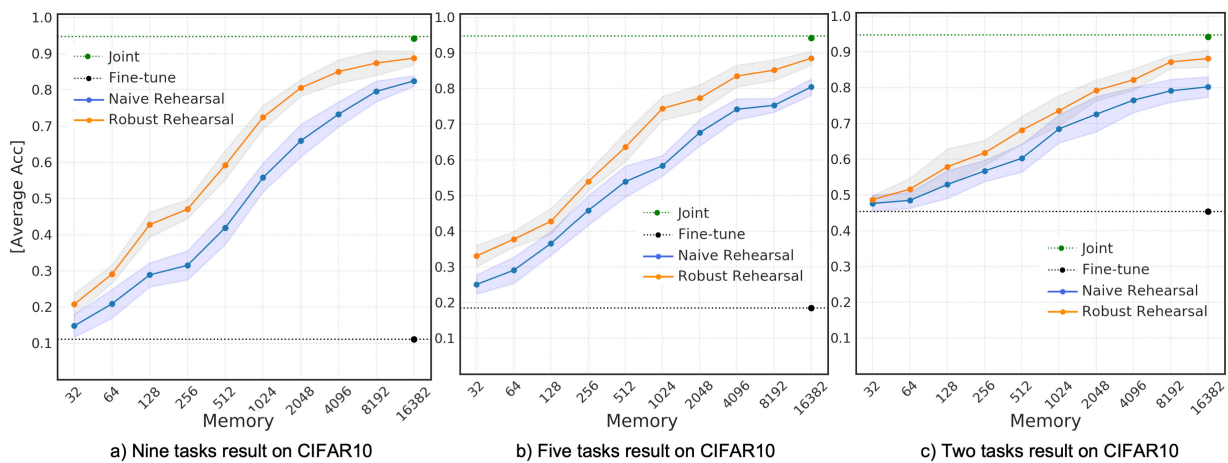

a) Nine tasks result on CIFAR10   b) Five tasks result on CIFAR10   c) Two tasks result on CIFAR10

FIGURE 7. Average accuracies from ten distinct experiments for both robust rehearsal and naive rehearsal strategies, compared across various memory sizes for nine, five, and two tasks on the CIFAR10 dataset, are presented. Sub-figures (a), (b), and (c) present average accuracies of the joint, fine-tuned, naive rehearsal, and robust rehearsal across varying buffer sizes for nine, five, and two tasks. The upper and lower dotted lines represent the average accuracies of the joint (i.e., upper bound) and fine-tune (i.e., lower bound) models, respectively. The robust rehearsal strategy's consistent outperforming of the naive rehearsal across varying memory sizes underscores its effectiveness in rehearsing CL-robust samples distilled via feature distillation and re-consolidation.

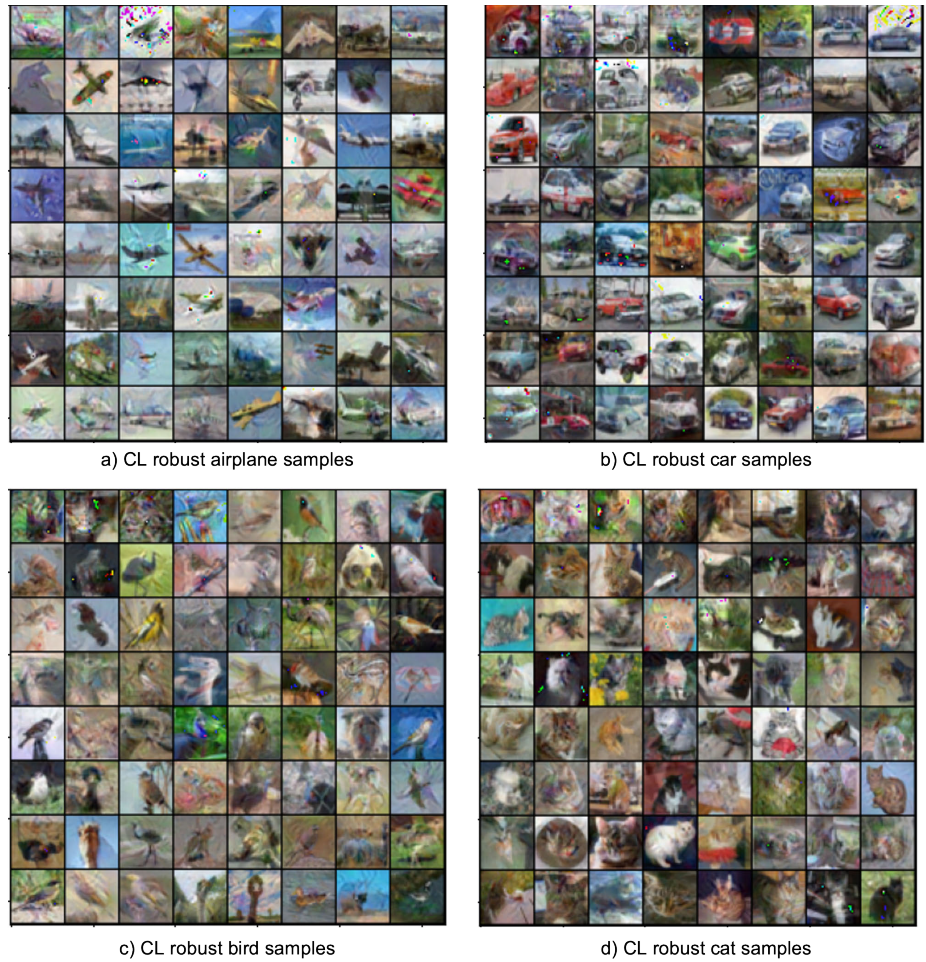

**FIGURE 8.** Sub-figures (a), (b), (c), and (d) visualize the CL-robustified features for the airplane, car, bird, and cat classes of the CIFAR-10 dataset, distilled through feature distillation. These CL-robustified features are rendered in a dream-like aesthetic that abstracts non-essential details while emphasizing each class's fundamental and distinctive attributes. The CL-robustified features contain only abstract dream-like minimum features of the class. The rehearsal of the minimum CL-robustified helps the CL model not to forget the previously learned classes. The CL-robustified features are characterized by dream-like minimal representations of each class. The rehearsal of CL-robustified features aids the CL model in retaining knowledge of previously learned classes.

### 4) VISUALIZING THE SPECIALIZED FEATURES UNDERSCORED THE VULNERABILITIES OF CL APPROACHES AGAINST ADVERSARIAL ATTACKS

A large body of literature underscores the vulnerability of existing CL approaches to various forms of attacks, such as backdoor attacks, adversarial attacks, and false memory formation [117], [118]. As Figure 6 illustrated, the training objectives influence the feature specialization of the underlying multi-task, continual, and adversarially robust model. Intriguingly, the visual resemblance of the CL-robust samples to the adversarially compromised images [48], as depicted in Figures 6, 2 and 3 suggest that CL-robust features and adversarially non-robust features can coexist, which underscores the vulnerability of CL to various adversarial attacks [40], [41]. These observations align with a large body of literature that already underscored the vulnerability of existing CL approaches to various forms of attacks, such as backdoor attacks, adversarial attacks, and false memory formation [41], [117], [118], [119]. Our previous work revealed that a CL model trained on adversarially robust features outperforms a CL model trained on standard datasets when subjected to different types of natural image distortions and adversarial attacks [40]. These findings emphasize the pivotal role played by the features in CL, which not only mitigate catastrophic forgetting but also define the vulnerability of CL models [40]. Reflecting on our results, we believe that our research not only underscores the neuroscience-inspired approaches for CL but also reveals promising avenues for future exploration, which include equipping CL models to continuously learn adversarially and CL-robust features, thereby mitigating catastrophic forgetting and securing all learned tasks against attacks.

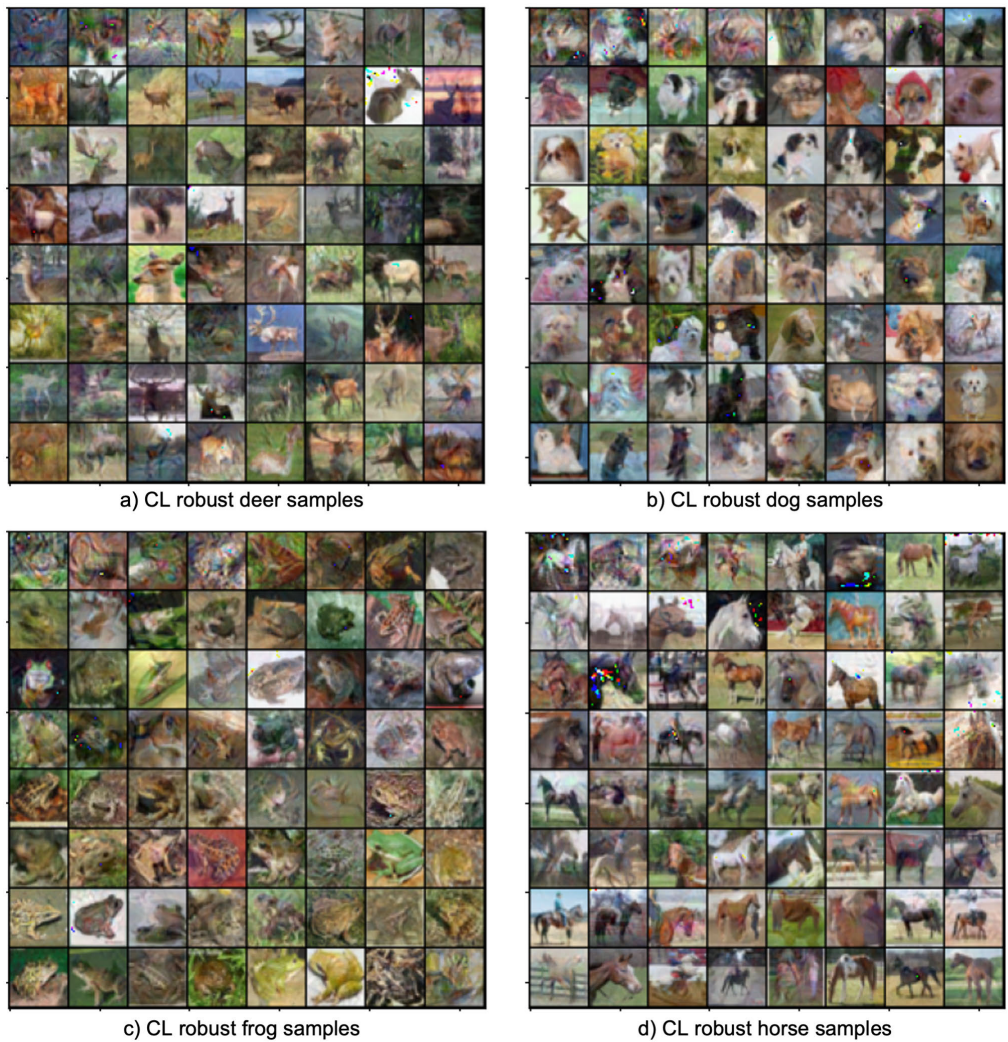

**FIGURE 9.** Sub-figures (a), (b), (c), and (d) visualize the CL-robustified features for the deer, dog, frog, and horse classes of the CIFAR-10 dataset, distilled through feature distillation. These CL-robustified features are rendered in a dream-like aesthetic that abstracts non-essential details while emphasizing each class's fundamental and distinctive attributes. The CL-robustified features contain only abstract dream-like minimum features of the class. The rehearsal of the minimum CL-robustified helps the CL model not to forget the previously learned classes. The CL-robustified features are characterized by dream-like minimal representations of each class. The rehearsal of CL-robustified features aids the CL model in retaining knowledge of previously learned classes.

### C. LIMITATIONS AND FUTURE WORK

A notable limitation is the requirement of additional computational time for feature distillation and re-consolidation to distill and re-distill the CL-robust samples, which must be performed before the model begins learning the next task. This limitation could be mitigated by incorporating self-attention in the model architecture, enabling robust feature distillation and re-consolidation during the learning phase itself [120]. The second limitation concerns using the Euclidean norm as the metric, which is a rudimentary distance metric that could be replaced with more sophisticated metrics, as indicated in the dataset distillation literature [34], [36]. Future research aims to expand the proposed solution and apply it to other varied tasks, including continual object segmentation. Moreover, the feature specialization underscored the vulnerability of CL to a range of attacks, such as backdoor, adversarial, and false memory formation attacks. These findings suggest promising future research directions, including the potential for equipping CL models to continuously learn adversarially, thereby reducing catastrophic

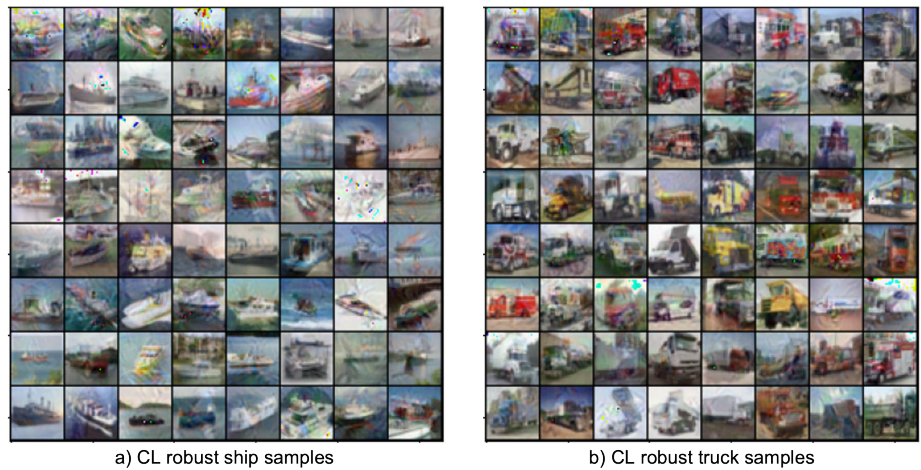

a) CL robust ship samples  b) CL robust truck samples

**FIGURE 10.** Sub-figures (a) and (b) visualize the CL-robustified features for the ship and truck classes of the CIFAR-10 dataset, distilled through feature distillation. These CL-robustified features are rendered in a dream-like aesthetic that abstracts non-essential details while emphasizing each class's fundamental and distinctive attributes. The CL-robustified features contain only abstract dream-like minimum features of the class. The rehearsal of the minimum CL-robustified helps the CL model not to forget the previously learned classes. The CL-robustified features are characterized by dream-like minimal representations of each class. The rehearsal of CL-robustified features aids the CL model in retaining knowledge of previously learned classes.

forgetting and enhancing the security of all learned tasks against such attacks.

## V. CONCLUSION

In this paper, we draw inspiration from neuroscience insights into memory consolidation and existing research in adversarial and continual learning, highlighting the crucial role of available features in shaping the model's overall performance. We developed a novel framework, Robust Rehearsal, which creates CL-robust memory samples that replicate the mammalian brain's memory consolidation and re-consolidation processes, i.e., rehearsing the distilled waking experiences to facilitate memory consolidation. Similarly, the proposed framework effectively distills and rehearses CL-robust samples while learning new tasks to retain the previously learned knowledge. We conducted extensive experiments on three datasets -CIFAR10, CIFAR100, and the real-world Helicopter Attitude datasets provided by the FAA- to demonstrate that CL models trained with robust rehearsal surpass standard baseline counterparts employing a traditional rehearsal-based strategy. The framework's superiority is further affirmed in scenarios with varying memory sizes and numbers of tasks, where it consistently outperforms counterpart approaches that do not utilize robust rehearsal. Additionally, our investigation into the impact of various optimization training objectives within joint, continual, and adversarial learning highlighted that the optimization objective in deep neural networks essentially dictates feature learning. Our findings suggest that closely adhering to neuroscience principles can substantially contribute to addressing the long-standing challenge of catastrophic forgetting in artificial intelligence systems.

## ACKNOWLEDGMENT
The authors are grateful to a Cliff Johnson Research Engineer, a Flight Test Engineer, and a Program Lead: Rotorcraft, Unmanned Aircraft, and eVTOL/Urban/Advanced Air Mobility, Manager (A) Data Analytics and CyberSecurity Section with the Software & Systems Branch, Aviation Research Division at the Federal Aviation Administration, for providing the Helicopter Attitude data.

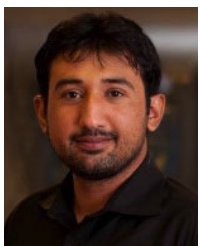

**HIKMAT KHAN** received the B.C.S. degree in computer science from Kohat University of Science and Technology (KUST), Pakistan, in 2010, and the M.S.C.S. degree in computer science from COMSATS University Islamabad (CUI), Pakistan, in 2017. He is currently pursuing the Ph.D. degree with Rowan University, Glassboro, USA. For his master's thesis, he focused on applying deep learning to brain tumor segmentation using MRI images. Before pursuing the master's degree, he was a Senior Software Engineer for over five years, in 2010. He was also a Machine Learning Engineer specializing in bio-medical imaging and natural language processing for over three years. During the Ph.D. degree, he was a Graduate Research Fellow, where he had the opportunity to work on multiple projects. Among these was a project supporting the Federal Aviation Administration (FAA) through a research grant/cooperative agreement, evaluating the feasibility of applying deep learning approaches to enhance safety within the rotorcraft industry. Concurrently with his Ph.D. studies, he was a Teaching Assistant for a year and a half and he was an Adjunct Faculty Member with Rowan University, teaching an introductory course in computer programming, in Fall 2023. He is also a part of the Rowan's Artificial Intelligence Laboratory (RAIL). His research was partially supported by the U.S. Department of Education through the Graduate Assistance in Areas of National Need (GAANN) Program under Award P200A180055. Moreover, he was an Entrepreneurial Lead with the esteemed National Science Foundation's (NSF) I-Corps Hub: Northeast Region Propelus Program, in May 2023, where he led the innovative Sepsis Early Watch Project, an Artificial Intelligence-Based Healthcare Initiative. His research interests include deep learning, continual learning, neuroscience, adversarial learning, and optimization.

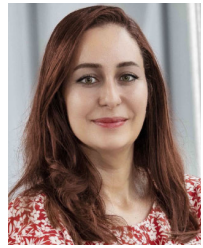

**NIDHAL CARLA BOUAYNAYA** (Member, IEEE) received the M.S. degree in pure mathematics from the University of Illinois at Chicago, in 2007, and the Ph.D. degree in electrical and computer engineering (ECE). She is currently a Professor in ECE and the Director of the Machine, Artificial Intelligence, and Virtual Reality Center (MAVRC), Rowan University. She is also the Associate Dean for Research and Graduate Studies with the Henry M. Rowan College of Engineering. Her research interests include big data analytics, machine learning, artificial intelligence, and mathematical optimization. She has coauthored over 120 refereed journal articles, book chapters, and conference proceedings, such as IEEE TRANSACTIONS ON PATTERN ANALYSIS AND MACHINE INTELLIGENCE, IEEE SIGNAL PROCESSING LETTERS, *IEEE Signal Processing Magazine*, and *PLOS Medicine*. She won numerous Best Paper Awards, the most recent was at the 2019 IEEE International Workshop on Machine Learning for Signal Processing. She also won the Top Algorithm at the 2016 Multinomial Brain Tumor Segmentation Challenge (BRATS). Her research is primarily funded by the National Science Foundation, The National Institutes of Health, U.S. Army Picatinny Arsenal, U.S. Department of Agriculture, the Federal Aviation Administration, and Lockheed Martin Inc., among others. She is also interested in entrepreneurial endeavors. In 2017, she co-founded MRIMATH LLC, a start-up company that uses artificial intelligence to improve patient oncology outcomes and treatment responses. MRIMath was awarded NIH SBIR Phases I and II contracts.

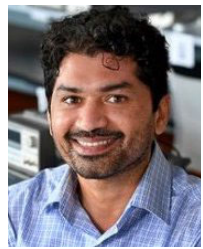

**GHULAM RASOOL** (Member, IEEE) received the B.S. degree in mechanical engineering from the National University of Sciences and Technology (NUST), Pakistan, in 2000, the M.S. degree in computer engineering from the Center for Advanced Studies in Engineering (CASE), Pakistan, in 2010, and the Ph.D. degree in systems engineering from the University of Arkansas at Little Rock, in 2014. He is currently an Assistant Member with the Department of Machine Learning, H. Lee Moffitt Cancer Center and Research Institute, Tampa, FL, USA. He was a Postdoctoral Fellow with the Rehabilitation Institute of Chicago and Northwestern University, from 2014 to 2016. Before joining Moffitt, he was an Assistant Professor with the Department of Electrical and Computer Engineering, Rowan University. His current research interests include building trustworthy multimodal machine learning and artificial intelligence models for cancer diagnosis, treatment planning, and risk assessment. Two National Science Foundation awards (NSF) funded his research efforts. Previously, his research was supported by the National Institute of Health (NIH), U.S. Department of Education, NSF, the New Jersey Health Foundation (NJHF), Google, NVIDIA, and Lockheed Martin Inc. His work on Bayesian machine learning won the Best Student Award at the 2019 IEEE Machine Learning for Signal Processing Workshop.